\pdfoutput=1
\documentclass[11pt]{article}

\usepackage[]{ACL2023}
\usepackage{times}
\usepackage{latexsym}

\usepackage[T1]{fontenc}
\usepackage[utf8]{inputenc}

\usepackage{microtype}

\usepackage{inconsolata}

\usepackage{graphicx}
\usepackage{enumerate}
\usepackage{amsfonts}
\usepackage{bm}
\usepackage{amsmath}
\usepackage{booktabs}
\usepackage{caption}
\usepackage{subcaption}
\usepackage{amssymb}
\usepackage{enumitem}
\usepackage{diagbox}
\usepackage{bbold}
\usepackage{multirow}
\usepackage[]{mdframed}
\usepackage{cleveref}

\newcommand{\ie}{\emph{i.e., }}
\newcommand{\eg}{\emph{e.g., }}

\newcommand{\cf}{\emph{cf. }}

\title{Think Twice Before Trusting: Self-Detection for Large Language Models through Comprehensive Answer Reflection}

\author{
Moxin Li\textsuperscript{1}, 
~Wenjie Wang\textsuperscript{1}$\thanks{$^{*}$Corresponding author.}$,
~Fuli Feng\textsuperscript{2, 3}\footnotemark[1], ~\textbf{Fengbin Zhu\textsuperscript{1}}, ~\textbf{Qifan Wang}\textsuperscript{4}\\
~\textbf{Tat-Seng Chua\textsuperscript{1}}
\\
\textsuperscript{1}National University of Singapore, 
~\textsuperscript{2}University of Science and Technology of China\\
\textsuperscript{3}Institute of Dataspace, Hefei, Anhui, China, 
~\textsuperscript{4}Meta AI
\\
\tt{limoxin@u.nus.edu}
,~\tt{wangwenjie@u.nus.edu},~\tt{fulifeng93@gmail.com}, \\
\tt{fengbin@nus.edu.sg},~\tt{wqfcr@meta.com}, ~\tt{dcscts@nus.edu.sg}\\
}

\begin{document}
\maketitle
\begin{abstract}

Self-detection for Large Language Models (LLMs) seeks to evaluate the trustworthiness of the LLM's output by leveraging its own capabilities, thereby alleviating the issue of output hallucination.
However, existing self-detection approaches only retrospectively evaluate answers generated by LLM, typically leading to the over-trust in incorrectly generated answers. 
To tackle this limitation, we propose a novel self-detection paradigm that considers the comprehensive answer space beyond LLM-generated answers. It thoroughly compares the trustworthiness of multiple candidate answers to mitigate the over-trust in LLM-generated incorrect answers. 
Building upon this paradigm, we introduce a two-step framework, which firstly instructs LLM to reflect and provide justifications for each candidate answer, and then aggregates the justifications for comprehensive target answer evaluation. 
This framework can be seamlessly integrated with existing approaches for superior self-detection. 
Extensive experiments on six datasets spanning three tasks demonstrate the effectiveness of the proposed framework.

%% 
%% 1. LLM vs LLMs 是不是统一下带不带s？目前看是大多数不带s的 
% 感觉都可以，统一就好。不带s的话当做单数来处理？加s啥的
%% 2. Related work是不是放到intro后或者conclusion前？
% 取决于related work对理解我们的formulation、method、exp有没有帮助，如果有的话 放到相应的前面，如果没有就放conclusion前面吧
%% 3. confidence score 应该是处于0 - 1 之间的，在图里是有大于1的值的，最高到1.4，是不是可以考虑限定在 0 - 1之间比较好？ 
% 看怎么处理，我看好像是直接截的？直接截的话不太好吧？
%% 4. evaluation vs explaination ?
% 这个是啥意思？
%% 5. 
%% 
%%

\end{abstract}

\section{Introduction}

Large Language Model (LLM) typically suffers from the hallucination issue, \cite{zhang2023siren, li2023halueval, golovneva2022roscoe, bang2023multitask}, 
which significantly undermines the trustworthiness of LLM's outputs. 
A promising research direction for evaluating the output trustworthiness and identifying incorrect outputs is self-detection \cite{zhao2023knowing, miao2023selfcheck, manakul-etal-2023-selfcheckgpt}. 
Given a question, self-detection aims to leverage LLM's own ability to evaluate the trustworthiness of its generated answers, without relying on external knowledge sources or specifically trained detection models. 
This paper investigates self-detection methods tailored for black-box API LLMs due to their excellent performance and the inherent challenge posed by limited output information \cite{achiam2023gpt, chatgpt}.

\begin{figure}[]
\setlength{\abovecaptionskip}{0.1cm}
\setlength{\belowcaptionskip}{0cm}
    \centering
    \includegraphics[width = 0.5\textwidth]{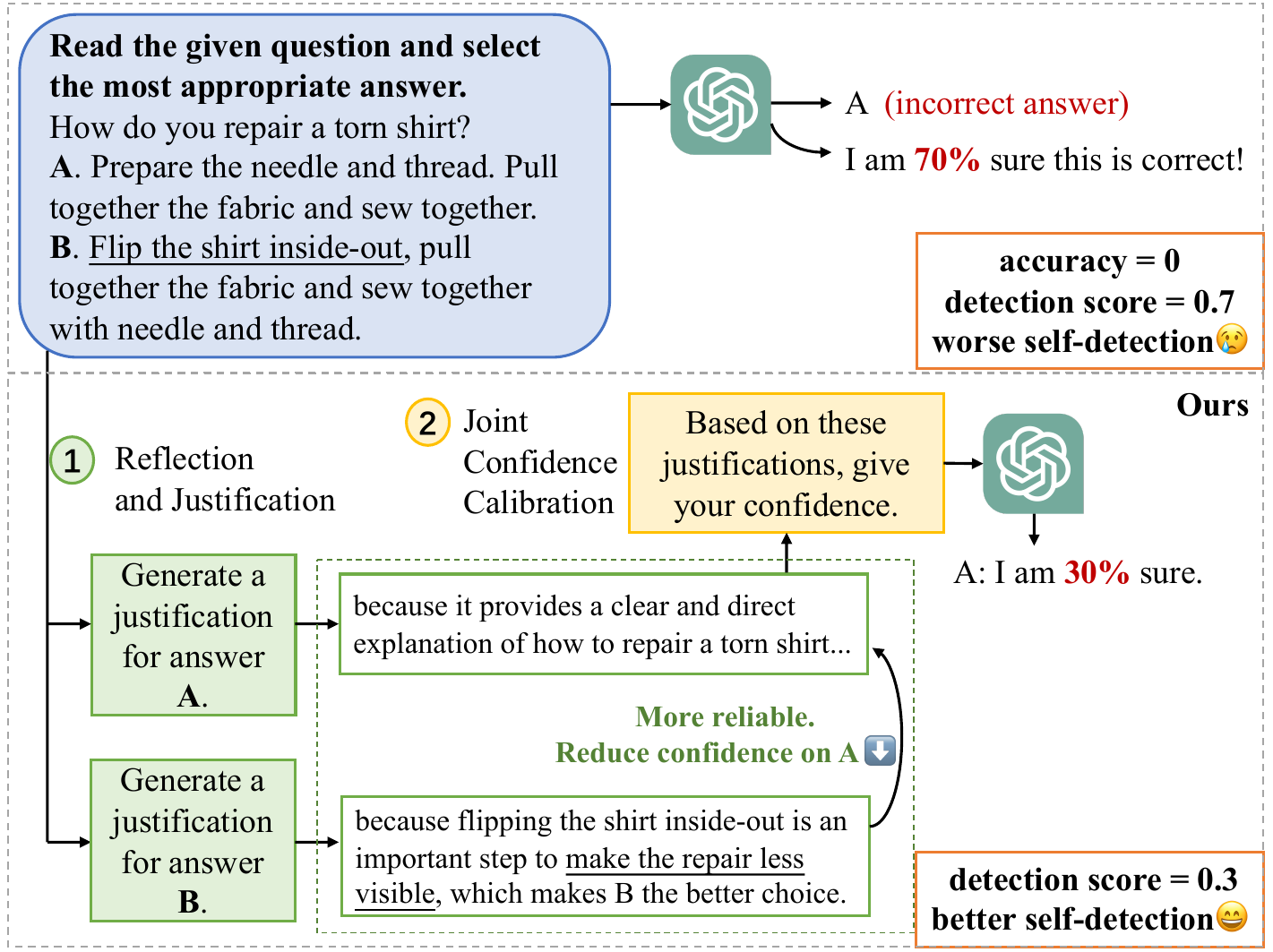}
    \caption{An illustration of \textit{Think Twice before Trusting} framework for mitigating the over-trust issue in LLM self-detection. LLM is instructed to reflect and generate justification on the trustworthiness of each answers before evaluating the trustworthiness of the target answer. }
    \vspace{-15pt}
    \label{fig:intro}
\end{figure}

Previous studies in self-detection can be broadly categorized into two paradigms (\cf Figure~\ref{fig:paradigms}). 
The first paradigm is confidence calibration, aiming to estimate LLM's confidence on the generated answer to align with the actual answer accuracy via multi-answer sampling and aggregation~\cite{xiong2023can, tian2023just, si2022prompting, jiang2023calibrating}. 
The second one is self-evaluation, which directly examines the compatibility of question and answer by designing various prompt strategies~\cite{miao2023selfcheck, kadavath2022language, weng-etal-2023-large}. 
These two paradigms have also been combined to enhance self-detection capabilities \cite{chen2023quantifying, ren2023self}.

However, both self-detection paradigms have shown a significant drawback: an inclination towards over-trusting the incorrect answers generated by LLM \cite{si2022prompting, xiong2023can, jiang2023calibrating, kadavath2022language}. 
We argue that one reason may be that
both paradigms merely evaluate LLM-generated answers, while LLM contains an inherent bias towards trusting its own generations~\cite{mielke2022reducing, lin2022teaching}, leading to serious over-trust in LLM-generated incorrect answers. 
An ideal self-detection paradigm should consider a more comprehensive answer space beyond LLM's generations. 
By evaluating on other potentially correct answers in a broader answer space, the strong validity in these answers can counterbalance the excessive trust in the incorrect LLM answers, thus alleviating the over-trust issue.

In this light, we introduce a new comprehensive answer evaluation paradigm involving the consideration of multiple candidate answers in the answer space to enhance self-detection (\cf Figure~\ref{fig:paradigms}). 
This paradigm meticulously evaluates each answer's trustworthiness as a correct answer to the question and aggregates these evaluations to enhance the self-detection of the target LLM answer.
The biased trust in the LLM-generated incorrect answers can be alleviated through the trustworthiness comparison with other more trustable answers. 
Our preliminary experiments reveal the efficacy of considering more comprehensive answers to confront over-trust (\cf Section~\ref{para:prelim_exp}). 
To summarize, two key considerations arise to instantiate this new paradigm: 1) resisting the inherent bias of LLM to precisely evaluate the trustworthiness of each question-answer pair, and 2) aggregating these evaluations 
in the trustworthiness evaluation of the target answer.

To this end, we present a novel self-detection framework to tackle the over-trust issue of LLMs, named \underline{T}hink \underline{T}wice before \underline{T}rusting ($T^3$) (\cf Figure~\ref{fig:intro}). 
Our framework pushes LLM to reflect and justify from different answers' perspectives before arriving at the trustworthiness on the target answer. 
Firstly, the LLM is instructed to generate justifications regarding the potential correctness of each answer. 
Subsequently, a prompt-based method is employed to integrate these justifications into joint evaluation for the target answer. 
Extensive experiments on six datasets across three tasks on three different LLMs show improved performance of $T^3$ over methods from existing paradigms. Notably, $T^3$ can be combined with existing methods for superior self-detection.
Our analysis also reveals $T^3$'s strong robustness and effective over-trust mitigation. 
Our contributions are three-fold. 
\begin{itemize}[leftmargin=*]
    \item We introduce a novel self-detection paradigm for mitigating the over-trust issue in LLM, addressing the limitation of existing paradigms by reflection in the broader answer space. 
    \item We present a novel $T^3$ framework to implement the comprehensive answer evaluation paradigm, which can be seamlessly integrated with existing self-detection methods. 
    \item We conduct extensive experiments on three NLP tasks with six datasets, validating the rationality and effectiveness of the proposed framework. 
\end{itemize}

\section{Problem Formulation}
\paragraph{LLM Self-Detection.}
We formulate the task of self-detection for LLM as follows. 
Given the input comprising of question $q$ combined with prompt $p$, which consists of an instruction and optional in-context examples, LLM can generate the answer $a$~\cite{brown2020language}, denoted as the target answer. 
Thereafter, self-detection aims to evaluate the trustworthiness of $a$ by LLM's own ability, generally in the form of a detection score $c \in \mathcal{R}$ \footnote{If the result of self-detection is a class label (trustable or untrustable), the detection score can be formulated as 1 or 0.}. 
The detection score $c$ can be used for indicating the actual accuracy of $a$, where low $c$ values indicate potentially incorrect answers. 
Denoting the self-detection strategy as a function $SD(\cdot)$, this process can be abstracted as 
\begin{align}
    \label{eq:llm_prediction}
    a & = LLM(p(q)),\\
    c & = SD(LLM(\cdot), q, a).
\end{align}

In the following, we illustrate the existing two paradigms for self-detection, \ie confidence calibration and self-evaluation, and introduce our proposed comprehensive answer evaluation paradigm. 

\begin{figure}[t]
\setlength{\abovecaptionskip}{0.1cm}
\setlength{\belowcaptionskip}{0cm}
    \centering
    \includegraphics[width = 0.47\textwidth]{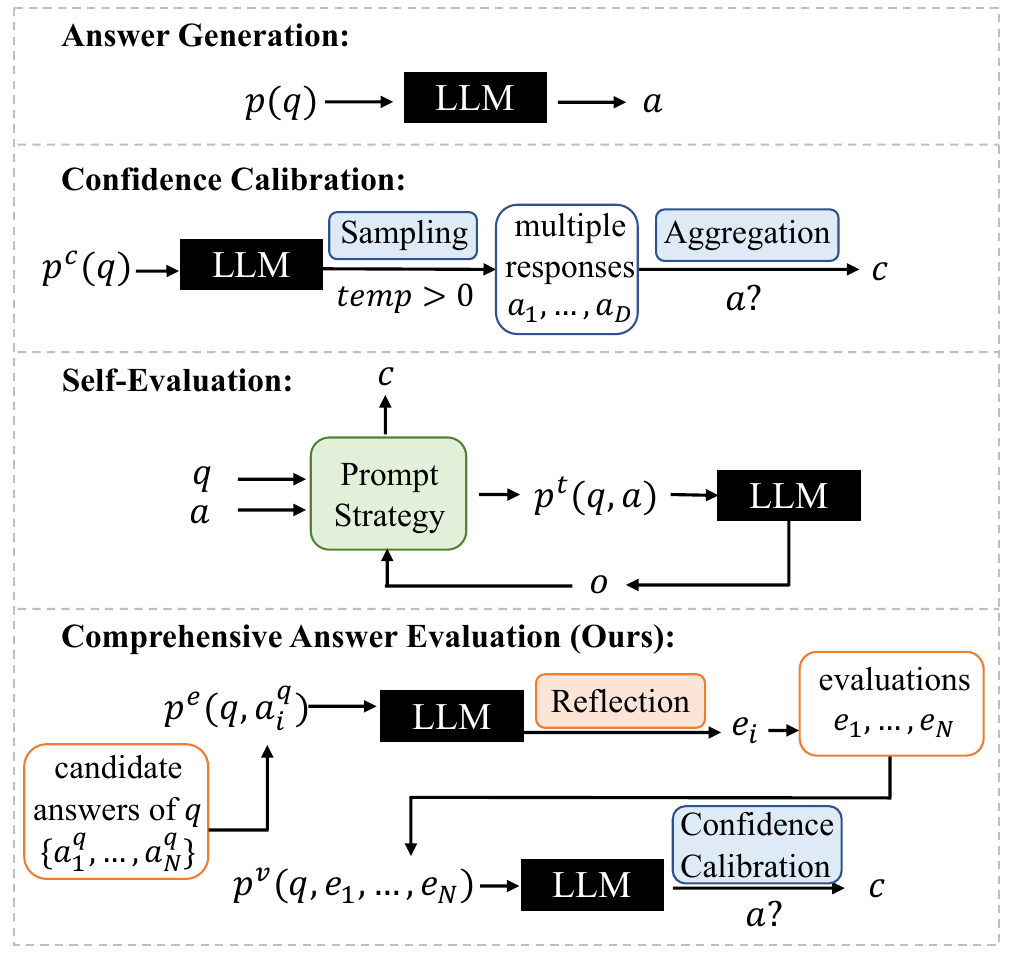}
    \caption{Two existing paradigms of self-detection and our new comprehensive answer evaluation paradigm. }
    \vspace{-10pt}
    \label{fig:paradigms}
\end{figure}

\paragraph{Confidence Calibration. } 
Confidence calibration aims to estimate LLM's level of certainty on the answer $a$, \eg estimating the LLM output probability of $a$, where the obtained confidence score as the detection score $c$ aims to calibrate with the actual answer accuracy. 
\citeauthor{xiong2023can} conclude a general three-step confidence calibration process for LLM as \textit{prompting, answer sampling, aggregation} (\cf Figure~\ref{fig:paradigms}). Denoting the prompt for confidence calibration as $p^c(\cdot)$ and the aggregation function as $Aggr(\cdot)$, this paradigm can be abstracted as, 
\begin{align}
    \label{eq:conf_calib}
    % & a_i = LLM(p^c(q)), \\ \notag
     c & = Aggr(a, \{a_1, ..., a_D\}), \\ \notag
     \text{where } a_i & = LLM(p^c(q)), i \in \{1, ..., D\}.
\end{align}
where $D > 1$ refers to the number of sampled answers. 
For example, self-consistency \cite{wang2022self, si2022prompting} aggregates the probability of $a$ in the sampled outputs of $p(q)$ (\eg using nucleus sampling \cite{Holtzman2020The}). Formally, 
\begin{align}
    \label{eq:self-cons}
     c&  = \frac{\sum_{i=1}^D \mathbb{1} (a_i = a)}{D}, \\ \notag
    \text{where }  a_i &= LLM(p(q)),  i \in \{1, ..., D\}. \notag
\end{align}
Besides, the Top-$K$ verbalized methods \cite{lin2022teaching, tian2023just} leverage a well-designed prompt $p^b$ (\cf Appendix~\ref{sec:appe_prompts}) to instruct the LLM to sample the $K$ most likely answers and output their corresponding probabilities in one response:  
\begin{align}
    \label{eq:verb}
    & [\{a_1, c_1\}, ... \{a_K, c_K\}] = LLM(p^b(q)). 
    % & c = c_i, \text{where } a = a_i, i \in \{1, ..., K\}. 
\end{align}
where $[\cdot]$ denotes the concatenation of the $K$ most likely answers with their probabilities. The probability of $a$ in the response is utilized as its detection score $c$ ($c=0$ if $a$ is not in the $K$ answers). 

However, confidence calibration methods are observed with severe over-trust issue on LLM, assigning high confidence score in some incorrectly generated answers \cite{si2022prompting, xiong2023can}. 
In fact, LLM has a bias to blindly trust its generated answers, leading to difficulties in distinguishing the correctness of its generated answers \cite{huang2023large, ling2023deductive, mielke2022reducing, ren2023investigating}. 
Although some attempts have been made to reduce high confidence in LLM and achieve better calibration \cite{jiang2023calibrating, zhao2024fact}, the over-trust issue still remains a severe problem towards effective self-detection. 

\paragraph{Self-Evaluation.} 
Self-evaluation methods concatenate $q$ and $a$ and leverage various designed prompts to instruct LLM in self-evaluating the correctness of $a$ from different perspectives. 
The prompt strategy examines the matching of $q, a$ by integrating the self-evaluation output $o$. 
Denoting the prompt strategy as a function $ES(\cdot)$, this process can be summarized as 
\begin{align}
    \label{eq:self-evaluation}
     c &= ES(q, a, o), \\
     \text{where } o &= LLM(p^t(q, a)). \notag
\end{align} 
where $p^t$ represents one prompt for self-evaluation. 

The shortcoming of self-evaluation is that many approaches under this paradigm are specifically designed for the mathematical question answering task, including step-wise checking on Chain-of-Thoughts (CoT) reasoning \cite{miao2023selfcheck}, completing masked $q$ using $a$ \cite{weng-etal-2023-large}, and natural program \cite{ling2023deductive}, limiting its applicability. The general method P(True) \cite{kadavath2022language} is straightforward and still demonstrates over-trust to incorrect LLM-generated answers. It directly asks LLM whether $a$ is the true answer to $q$ via the prompt $p^r$ (\cf Appendix~\ref{sec:appe_prompts}), and uses the probability of ``True'' in the sampled responses as $c$. Formally, 
\begin{align}
     c &= \frac{\sum_{i=1}^D \mathbb{1} (o_i = \textit{True})}{D}, \\ \notag
    \text{where }  o_i &= LLM(p^r(q, a)), i \in \{1, ..., D\}.
\end{align}
The two paradigms can be combined for better self-detection \cite{xiong2023can, chen2023quantifying, ren2023self, agrawal2023language}.

\paragraph{A New Comprehensive Answer Evaluation Paradigm. }
A notable limitation of the existing two paradigms is that their evaluation merely involves LLM-generated answers $a_i$, in which LLM may exhibit over-trust. 
We argue that such biased over-trust could be alleviated if LLM had thoroughly compared the trustworthiness of more candidate answers of $q$ beyond LLM-generated answers. 
We consider the multi-choice question answering setting where a comprehensive answer space is provided. \footnote{For other settings, the answer space can be obtained via answer retrieval or additional model prediction.}
If other answers in $q$'s answer space had a strong tendency to be correct, the high detection score for LLM-generated incorrect $a$ could be diminished, reducing the over-trust issue. 

In this light, we propose a novel comprehensive answer evaluation paradigm that considers $N$ potential answers in $q$'s answer space, denoted as $\{a^q_1, a^q_2, ..., a^q_N\}$. 
First, LLM evaluates the trustworthiness of each $(q, a^q_i)$ pair using the prompt $p^e$.
\begin{align}
     \label{eq:gen_expl} 
      &e_i = LLM(p^e(q, a^q_i)), i \in \{1, ..., N\}.  
\end{align}
Then, the obtained evaluations $e_1,...,e_N$ can be integrated into existing paradigms to derive a more refined detection score for $a$ by adjusting the prompts. For example, by adjusting the $p^c$ for confidence calibration (\cf Eq~\ref{eq:conf_calib}) into a new prompt $p^v$, the detection score for $a$ can be derived as, 
\begin{align}
     \label{eq:para_step_2} 
      c  &= Aggr(a, \{a_1, ..., a_i\}),  \\
      \text{where } a_i &= LLM(p^v(q, e_1, ..., e_N)), i \in \{1, ..., D\}. 
     \notag
\end{align}
The evaluations can also be integrated into self-evaluation by adjusting the prompt $p^t$ in Eq~\ref{eq:self-evaluation}. 
% This paradigm can also be combined with existing paradigms for better self-detection (\cf Section~\ref{sec:main_results}). 

\paragraph{Preliminary Experiments.} \label{para:prelim_exp}

\begin{figure}
    \centering
    \setlength{\abovecaptionskip}{0.1cm}
    \setlength{\belowcaptionskip}{0cm}
    \includegraphics[width = 0.5\textwidth]{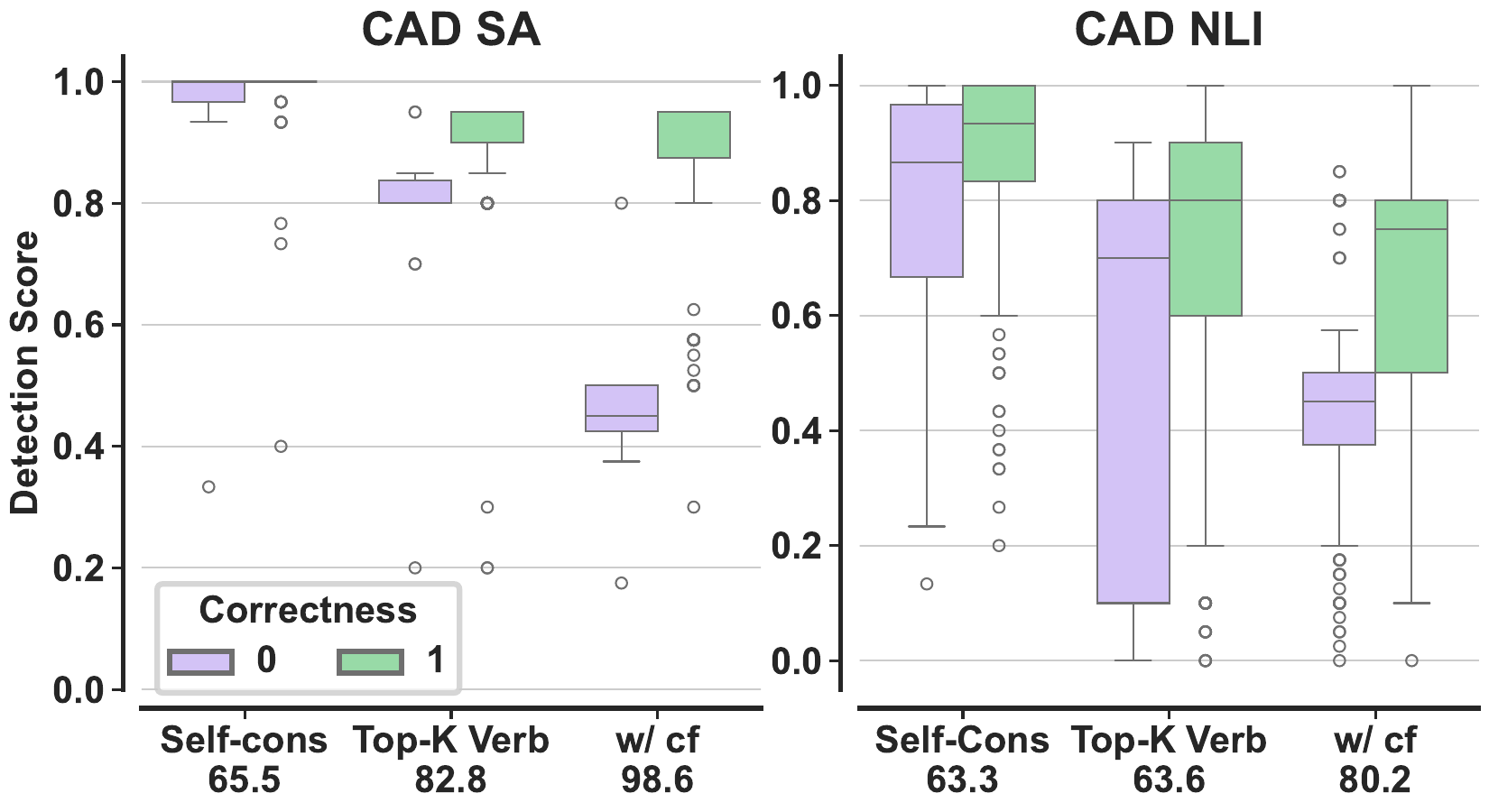}
    \caption{Comparison of self-detection methods on CAD. \textit{w/ cf} denotes our strategy with counterfactual data. The AUROC values are shown in the x-axis. The boxes on the left and right represent the detection scores of incorrect and correct instances, respectively. }
    \vspace{-8pt}
    \label{fig:prelim_cf}
\end{figure}

We conduct a preliminary experiment to validate that considering more answers in the answer space to adjust the detection score is beneficial for self-detection. 

Our hypothesis is that the evaluation of other answers can be leveraged to mitigate over-trust in the incorrect $a$. 
To demonstrate this, we employ counterfactual questions $\bar{q}$, which is minimally edited from $q$ to have a different label within $q$'s answer space. 
We aim to utilize the label difference between $q$ and $\bar{q}$ to identify unreliable LLM-generated answer for $q$ and adjust its detection score. 
Suppose the LLM-generated answers for $\bar{q}$ and $q$ are $\bar{a}$ and $a$, respectively. 
If $\bar{a}$ equals $a$, $a$ and $\bar{a}$ must have at least one wrong answer since $\bar{q}$ and $q$ have different labels. Thus the detection score of $a$ on $p(q)$ (denoted as $c_a$) should be reduced according to the detection score of $\bar{a}$ on $p(\bar{q})$ (denoted as $c_{\bar{a}}$) because the increasing of $c_{\bar{a}}$ indicates the weakened $c_a$. 
Conversely, if $\bar{a}$ differs from $a$, $a$ and $\bar{a}$ are relatively trustable, and $c_a$ can be an average of itself and $c_{\bar{a}}$. 
Formally, $c_a$ is re-calculated as 
\begin{align}
    c = 
    \begin{cases}
      \frac{1}{2} (c_a + c_{\bar{a}} )  & \text{if $a \neq \bar{a}$}, \\
      \frac{1}{2} (c_a + O(c_{\bar{a}}) )  & \text{else}.
    \end{cases}       
\end{align}
where $O(c_{\bar{a}})$ denotes the detection score that $\bar{q}$'s label is not $\bar{a}$. In a $k$-classification task, we roughly estimate $O(c_{\bar{a}}) = \frac{1}{k-1}(1 - c_{\bar{a}})$. 

We experiment with the CAD dataset \cite{kaushik2019learning}, which contains human-annotated original and counterfactual data pairs for sentiment analysis (SA) and natural language inference (NLI) tasks. 
We compare the AUROC with self-consistency and Top-$K$ verbalized methods to evaluate the self-detection performance on GPT-3.5 (see Section~\ref{sec:compared_methods} and Appendix~\ref{sec:appe_prelim_exp} for more details).

Figure~\ref{fig:prelim_cf} shows the AUROC and the statistics of detection scores for correct and incorrect $q, a$ instances, respectively. We can observe that 
1) the self-consistency and Top-$K$ verbalized methods have notable over-trust. The detection scores for incorrect instances have large overlap with the correct ones, making it challenging to distinguish them.
2) Our strategy, denoted as \textit{w/ cf}, improves AUROC by lowering detection scores on incorrect instances, showing that considering other answers can potentially alleviate the over-trust in incorrect answers. 
However, human-annotated counterfactual data is not easily available \cite{li-etal-2023-hypothetical}, motivating us to propose the following framework.

\section{Think Twice Before Assure Framework}
Implementing the proposed paradigm involves two key considerations. 
First, given the potential bias of LLM over-trust in the generated answer $a$, it is essential to develop strategies to resist this bias and thoroughly evaluate the trustworthiness of each answer $a^q_i$. 
Secondly, it is crucial to derive strategies to effectively combine these evaluations for effective self-detection of $a$. 
To address these concerns, we introduce the following two-step framework. 

\paragraph{Step 1: Reflection and Justification.}
We first instruct LLM to reflect on the trustworthiness of each answer $a^q_i$ and force LLM to seek justification for $a^q_i$ as the correct answer of $q$, as defined by Eq.~\ref{eq:gen_expl}. 
The LLM is instructed with the prompt $p^e$ in Table~\ref{tab:prompt_e_and_v} to gather comprehensive evidence $e_i$ from its knowledge, in order to support the rationality of using $a^q_i$ to answer $q$. 
The rationality of this step is that $p^e$ instructs LLM to abduct the justification from $q$ and $a^q_i$, which avoids the LLM bias that lies in the generation direction from $p(q)$ to $a$. As a minor clue, generating CoT explanations from $p(q)$ before $a$ has been validated to be ineffective for calibration \cite{zhang2023study}.

\begin{table}[h]
    \centering
    \resizebox{0.48\textwidth}{!}{
    \begin{tabular}{l|p{8.2cm}} 
    \toprule
    \multirow{3}{*}{$p^e$} & The task is to [task description]. \\
    &Question: [$q$]. Answer choices: [$a^q_1, ..., a^q_N$]. \\
    &The answer is [$a^q_i$]. \\
    &Please generate an explanation to try to justify the answer judgment. \\  
    \midrule
    \multirow{6}{*}{$p^v$} &The task is to [task description]. \\
    &Provide your $N$ best guesses and the probability that each is correct (0.0 to 1.0) for the following question... \\ 
    &Question: [$q$]. Answer choices: [$a^q_1, ..., a^q_N$]. \\
    &Possible explanation 1: [$e^1$]  \\
    &... \\
    &Possible explanation $n$: [$e^N$]\\
    \bottomrule
    \end{tabular}
    }
    \setlength{\abovecaptionskip}{0.1cm}
\setlength{\belowcaptionskip}{0cm}
    \caption{Prompts used in our $T^3$ framework. $p^e$ prompts LLM to reflect and generate justification $e_i$ for each $a^q_i$, and $p^v$ prompts LLM to estimate confidence according to different $e_i$.}
    \label{tab:prompt_e_and_v}
    \vspace{-10pt}
\end{table}

\paragraph{Step 2: Joint Confidence Calibration.}  \label{para:joint_confidence_calibration}
After obtaining the justification $e_i$ for each $a^q_i$, we choose to integrate these $e_i$ with a confidence calibration method, the Top-$K$ verbalized (\cf Eq.~\ref{eq:verb}) to derive the confidence of answer $a$ as the detection score.
We choose this method due to its capability to generate a set of $K$ potential answers and their respective probabilities efficiently in a single response, where we set $K$ as the number of answers $N$. 
As indicated in the prompt $p^v$ of Table~\ref{tab:prompt_e_and_v}, the generated justifications $e_i$ can be seamlessly integrated for confidence calibration of Top-$K$ verbalized.

An alternative approach to determine the final detection score is to put one justification to each $p^v$, generating $N$ distinct confidence scores for answer $a$, and then compute the averaged confidence score as the detection score.
%\begin{align} \label{eq:woavg}
%    c & = \frac{1}{N} \sum_{i=1}^N Pr(LLM(\cdot), p^v(q, e_i), a)
%\end{align}
We do not choose this setting as prompting LLM to estimate from different perspectives via a unified prompt is more efficient and effective than a simple average of the confidence scores (further validated in Section~\ref{sec:ablation_studies}). 
Moreover, we find that the detection scores are sensitive to the order of justification in $p^v$, thus we shuffle the order of $e^i$ in $p^v$ and compute the averaged score. 
Notably, the $T^3$ framework can be combined with existing approaches, such as prompt ensemble \cite{jiang2023calibrating}, and Hybrid method which adjust the detection score based on the difference with other methods \cite{xiong2023can}.

\section{Related Work}

\paragraph{Confidence Calibration of LLM.}

Confidence calibration has been previously studies in neural networks \cite{guo2017calibration} and applied in NLP models \cite{desai2020calibration, dan2021effects, hu2023uncertainty}. 
After the advent of LLM, many confidence calibration methods utilize the output token probability, such as semantic uncertainty \cite{kuhn2023semantic}, temperature scaling \cite{shih2023long}, entropy-based \cite{huang2023look}, semantic significance \cite{duan2023shifting}, and fine-tuning for calibration \cite{jiang2021can, lin2022teaching}. \citet{zhang2023sac} also employ model ensemble for calibration. Our research is orthogonal to them, since we focus on black-box API LLM itself. Other recent work suitable for black-box LLM includes fidelity elicitation \cite{zhang2024calibrating}, fact elicitation \cite{zhao2024fact} and perturbation generation \cite{gao2024spuq}. 

% To tackle the overconfidence issue, the first category of methods also includes 
%question rephrasing \cite{zhao2023knowing}, instruction and prompt template ensemble \cite{jiang2023calibrating}, 
% answer choice shuffling \cite{ren2023self}, and reflection from multiple perspective \cite{zhang2024self}. \citet{zhang2023sac} also employ model ensemble for better calibration. 
%The second category of method also includes program-like evaluation on CoT \cite{ling2023deductive}, %miao2023selfcheck}, 
%generating and executing verification codes \cite{zhou2023solving}, 
%asking verification questions \cite{manakul-etal-2023-selfcheckgpt}, while some of them are limited to certain domains. 
% Notably, the Top-$K$ verbalized \cite{tian2023just}, the self-consistency \cite{si2022prompting}, and their Hybrid \cite{xiong2023can} methods also involve the confidence of other answers, yet the estimation of their confidences is also affected by the LLM bias and thus these answers do not genuinely contribute to the overconfidence mitigation of the target answer.

\paragraph{Self-Evaluation of LLM.}
LLM self-evaluation often focuses on specific domains, \eg code generation \cite{zhou2023solving}, natural language generation \cite{lin2023generating} and fact checking \cite{manakul-etal-2023-selfcheckgpt}. The general methods include P(True)~\cite{kadavath2022language} and directly asking LLM \cite{li2024confidence}. 
\citet{feng2024don} also performs answer reflection and employs model collaboration, yet they still focus on answers generated by LLM. 
Note that self-detect \cite{zhao2023knowing} is also a general self-evaluation method following the three-step confidence calibration pipeline. 

Other works that are related but orthogonal to us include training independent models for LLM evaluation \cite{wang2023learning, li2023making, grace2023, zhao2023automatic, li2024inference}, and using external tools for LLM verification \cite{min-etal-2023-factscore, ni2023lever}. They are usually applied to specific domains, while we aim at LLM self-detection for general tasks. Also, fine-tuning LLM for better trustworthiness \cite{an2023learning, tian2023fine} is orthogonal to us. 

\paragraph{Application of LLM Self-Detection.}
The outcome of self-detection can be applied in many ways to avoid hallucination and erroneous outputs, such as identifying potentially hallucinated generation for knowledge retrieval and verification \cite{zhao-etal-2023-verify}, guided output decoding \cite{xie2023self}, identifying ambiguous questions \cite{hou2023decomposing}, selective generation \cite{ren2023self, zablotskaia2023uncertainty}, and LLM self-improve \cite{huang-etal-2023-large}. More applications can be found in this survey \cite{pan2023automatically}. 
% reflection from multiple perspective \cite{zhang2024self}. 这篇没分数。这种self-refine都没有分数不算。

\paragraph{Generating Multiple-Side Rationale.}
Some existing research on LLM also utilizes multiple-side rationales similar to $T^3$, such as Process of Elimination \cite{balepur2023s}, Maieutic Prompting \cite{jung2022maieutic}, and Debate \cite{michael2023debate}. 
% \citet{balepur2023s} find that LLM performs worse in Process of Elimination than direct question answering, and faces consistency issues. 
%\citet{jung2022maieutic} propose Maieutic Prompting for binary QA tasks which jointly reasons over abductive and recursive rationales.
%\citet{liu2023score} analyze the contradictions between rationales and answers and test LLM’s ability in recognizing such contradictions.
%\citet{michael2023debate} use the debate of two unreliable LLM experts to help a non-expert human to judge the answer. 
Our $T^3$ differs from these work mainly in that $T^3$ incorporates multiple side rationales into LLM self-detection, by combining the rationales with Top-$K$ Verbalized confidence estimation with designed prompts. 
Besides, the format of multi-side rationale of $T^3$ is only similar to \citet{jung2022maieutic}, while $T^3$ can handle a wider range of tasks.

\section{Experiments}

\paragraph{Setup.}
We conduct experiments on six datasets across three tasks. 
IMDB \cite{maas-EtAl:2011:ACL-HLT2011} and Flipkart \cite{flipkart} for SA, SNLI \cite{bowman-etal-2015-large} and HANS \cite{mccoy-etal-2019-right} for NLI, CommonsenseQA \cite{talmor2019commonsenseqa} and PIQA \cite{bisk2020piqa} for commonsense question answering (CQA).
For LLMs, we utilize GPT-3.5 (\textit{gpt-3.5-turbo-1106}) from OpenAI\footnote{\url{https://openai.com/blog/openai-api}.}, GLM-4 \cite{du2022glm} from ZhipuAI\footnote{\url{https://open.bigmodel.cn/}.}, and Gemini (\textit{gemini-1.0-pro-001}) from Google\footnote{\url{https://gemini.google.com/app}.}.
% which are among the most powerful LLMs. 
Dataset statistics and LLM hyperparameters are listed in \Cref{sec:appe_llm_param,sec:appe_dataset}. 

% GPT-4 (\textit{gpt-4-0613}) 

\paragraph{Compared Methods.} \label{sec:compared_methods}
We utilize the following categories of compared methods. 
For the first paradigm, we include
\textbf{Self-cons} \cite{wang2022self} (\cf Eq.~\ref{eq:self-cons}), 
\textbf{CoT-cons}, an extension of Self-cons by instructing LLM to output the CoT reasoning before the answer, 
\textbf{Induced-cons} \cite{xiong2023can}, adding different misleading contexts to the input to examine consistency. 
\textbf{Top-$K$ Verb} \cite{tian2023just} (\cf Eq.~\ref{eq:verb}), 
\textbf{Hybrid} \cite{xiong2023can}, an integration of Top-$K$ Verb and Self-cons/CoT-cons, where we show the better results, 
\textbf{Self-detect} \cite{zhao2023knowing}, taking the answer entropy of multiple rephrased questions, 
and \textbf{CAPE} \cite{jiang2023calibrating}, a prompt ensemble method that we implement on Top-$K$ Verb. 
For the second paradigm, we utilize the general 
\textbf{P(True)} \cite{kadavath2022language}, and 
\textbf{Self-Probing} \cite{xiong2023can}, directly asking LLM to explain and evaluate the confidence of the given answer. 
Finally, to show the flexibility of $T^3$ in combining with existing methods to further improve self-detection, we show the performance of Hybrid $T^3$ with Top-$K$ Verb (\textbf{$T^3$ + Top-$K$ Verb}), and $T^3$ with prompt ensemble following CAPE (\textbf{$T^3$ + PE}). 
For a fair comparison, we generate the target answer for each dataset with LLM temperature as 0, and compare all methods based on this target answer (\cf Eq~\ref{eq:llm_prediction}).
More details are in \Cref{sec:appe_add_detail,sec:appe_prompts}, including a comparison on the number of API calls showing $T^3$'s reasonable cost. 

% Additional details, the number of API calls, and the prompts can be found in \Cref{sec:appe_add_detail,sec:appe_prompts}.
\paragraph{Evaluation Metrics.}
We mainly use \textbf{AUROC} \cite{boyd2013area} and \textbf{PRAUC} \cite{manning1999foundations} to evaluate the self-detection ability.
%, both ranging from 0 to 1. 
They assess the effectiveness of detection scores in distinguishing answer correctness using true positive/false positive and precision/recall curves, respectively.
% 为什么比ECE？
Additionally, we use the Expected Calibration Error (\textbf{ECE}) to evaluate the calibration performance for confidence calibration methods. 
 
\subsection{Results} \label{sec:main_results}

\begin{table}[hbt!]
    \begin{subtable}[h]{0.5\textwidth}
        \centering
        \resizebox{\textwidth}{!}{
\begin{tabular}{lllllll}
\toprule
      & \multicolumn{3}{l}{IMDB} & \multicolumn{3}{l}{Flipkart} \\
    \cmidrule{2-4}  \cmidrule{5-7} & AUROC $\uparrow$       & PRAUC $\uparrow$ & ECE $\downarrow$ & AUROC $\uparrow$       & PRAUC $\uparrow$ & ECE $\downarrow$   \\
\midrule
Self-cons & 65.5 & 96.8 & 0.115 & 71.4 & 91.4 & 0.106\\
CoT-cons & 75.6 & 97.7 & 0.104 & 72.8 & 91.9 & 0.133\\
%Induced-cons & & & & 64.5 & 89.2 & \\
Top-$K$ Verb & 82.8 & 98.5 & 0.242 &79.3 & 93.7 & 0.131 \\
P(True) & 80.1 & 98.1 & 0.104 & 54.5 & 86.7 & 0.243\\
%Self-Probe & & & & 72.8 & 92.3 & \\
Hybrid & 87.0 & \underline{98.8} & 0.183 & 79.5 & 94.2 & 0.176 \\
Self-detect& 68.9  & 97.1 & 0.320& 71.2 & 91.4&0.146\\
CAPE& 87.7  & \textbf{98.9} & 0.096 & 76.4 & 93.9&0.107\\
\midrule
$T^3$ & 87.9  & \textbf{98.9} & \textbf{0.045} & \underline{81.3} & \underline{94.5}& \textbf{0.015}\\
\ \small{+ Top-$K$ Verb} & \underline{88.0}& \textbf{98.9}& \underline{0.052} & \textbf{81.6} & \textbf{94.9}& \underline{0.019}\\
\ \small{+ PE} & \textbf{88.1}& \textbf{98.9}& 0.056 & 74.2& 92.9& 0.157\\
\bottomrule
\end{tabular}
       }
       \caption{SA. }
       \label{tab:main_sa}
    \end{subtable}
    % \hfill 
    \\
    \begin{subtable}[h]{0.5\textwidth}
        \centering
        \resizebox{\textwidth}{!}{
\begin{tabular}{lllllll}
\toprule
                & \multicolumn{3}{l}{SNLI}             & \multicolumn{3}{l}{HANS}            \\
\cmidrule{2-4}  \cmidrule{5-7}& AUROC $\uparrow$       & PRAUC $\uparrow$ & ECE $\downarrow$ & AUROC $\uparrow$       & PRAUC $\uparrow$ & ECE $\downarrow$\\
\midrule
Self-cons & 63.3 & 71.4 &0.047& 56.0 & 64.8 &0.051\\
CoT-cons & 66.7 & 73.8 &0.043& 59.4 & 67.9& 0.152\\
%Induced-cons & & & &63.4 &70.1 & \\
Top-$K$ Verb & 63.6 & 74.0 &0.089& 53.3 & 64.9 &0.273\\
P(True) & 55.4 & 67.4 &0.117& 60.8 & 70.1 &0.067\\
%Self-Probe & & & & 58.2 & 66.1 & \\
Hybrid & 66.7 & 78.8 &\underline{0.029}& 62.0 & 71.1 &0.193\\
Self-detect& 59.3  & 68.5 &0.142& 55.3 & 64.5 &0.063\\
CAPE& 69.0 & 79.6 &0.030& \underline{71.9} & \underline{80.1}&\underline{0.028}\\
\midrule
$T^3$ & \textbf{77.9}  & \underline{84.6}  &0.157& 69.9    & 77.5 & \textbf{0.022}\\
\ \small{+ Top-$K$ Verb} & \underline{77.1}  & \textbf{84.7} &\textbf{0.024}& 71.3  & 79.6  & 0.030\\
\ \small{+ PE} & 70.8  & 76.7  &0.130& \textbf{74.5}  & \textbf{81.2} & 0.034 \\
\bottomrule
\end{tabular}
    }
       \caption{NLI.}
       \label{tab:main_nli}
    \end{subtable}
    % \hfill
    \\
    \begin{subtable}[h]{0.5\textwidth}
        \centering
        \resizebox{\textwidth}{!}{
\begin{tabular}{lllllll}
\toprule
             & \multicolumn{3}{l}{CommonsenseQA}                & \multicolumn{3}{l}{PIQA}                  \\
\cmidrule{2-4}  \cmidrule{5-7}& AUROC $\uparrow$       & PRAUC $\uparrow$ & ECE $\downarrow$& AUROC $\uparrow$       & PRAUC $\uparrow$ & ECE $\downarrow$\\
\midrule
Self-cons & 70.7 & 81.7 &0.151& 78.6 & 94.0 &0.043\\
CoT-cons & 81.8 & 88.9 &0.049& 76.7 & 94.2& 0.097\\
%Induced-cons & & & &71.4 & 81.6& \\
Top-$K$ Verb & 69.4 & 81.5 &0.026& 76.8 & 93.3&0.060 \\
P(True) & 62.5 & 78.0 &0.097& 71.9 & 93.9&0.176 \\
%Self-Probe & & & & 58.6& 75.1& \\
Hybrid & 77.5 & 89.0 &\underline{0.015}& 82.4 & 95.5 &0.088\\
Self-detect& 67.9 & 81.5 &0.261& 68.5 & 91.0&0.161\\
CAPE& 78.7  & 88.8 &0.021& \underline{87.9} & \underline{97.8}&0.067\\
\midrule
$T^3$ & 83.5  & 90.7  &\textbf{0.009}& 83.4 & 95.2 & \underline{0.016}\\
\ \small{+ Top-$K$ Verb} & \textbf{85.8} & \textbf{93.4} &0.017& 85.3 & 96.2 &\textbf{0.010}\\
\ \small{+ PE} &  \underline{84.4}& \underline{92.1} &0.019& \textbf{90.3} & \textbf{97.9}&0.034 \\
\bottomrule
\end{tabular} 
        }
        \caption{CQA. }
        \label{tab:main_csqa}
     \end{subtable}
    \setlength{\abovecaptionskip}{0.1cm}
    \setlength{\belowcaptionskip}{0cm}
     \caption{Results of the compared methods on GPT-3.5. Bold font and underline indicate the best and second best performance, respectively.  }
     \vspace{-10pt}
     \label{tab:main}
\end{table}

\begin{table*}[t!]
\centering
\resizebox{\textwidth}{!}{
\begin{tabular}{lllllllllllll}
\toprule
      & \multicolumn{2}{l}{IMDB} & \multicolumn{2}{l}{Flipkart} & \multicolumn{2}{l}{SNLI}& \multicolumn{2}{l}{HANS} & \multicolumn{2}{l}{CommonsenseQA} & \multicolumn{2}{l}{PIQA} \\
    \cmidrule{2-3}  \cmidrule{4-5} \cmidrule{6-7} \cmidrule{8-9}  \cmidrule{10-11} \cmidrule{12-13}& AUROC$\uparrow$ & PRAUC$\uparrow$ & AUROC$\uparrow$ & PRAUC$\uparrow$  & AUROC $\uparrow$ & PRAUC $\uparrow$& AUROC$\uparrow$ & PRAUC$\uparrow$ & AUROC$\uparrow$ & PRAUC$\uparrow$  & AUROC $\uparrow$ & PRAUC $\uparrow$ \\
\midrule
$T^3$ & \textbf{87.9}  & \textbf{98.9} & \textbf{81.3} & \textbf{94.5}& \textbf{77.9} & \textbf{84.6} & \textbf{69.9} & \textbf{77.5} & \textbf{83.5} & 90.7 & 83.4 & 95.2  \\
w/ CoT expl& 72.4 & 97.5 & 76.6 & 93.4 & 67.1 & 75.2& 53.7 & 64.1 & 78.7 & 86.8 & 81.3 & 94.8 \\
sep expl & 86.5 & 98.8& 79.5& 94.2& 68.5& 75.3& 54.1 & 63.8 & 80.3 & 87.8 &\textbf{84.0} & \textbf{95.8} \\
w/o shuffle& 75.9  & 98.3 & 71.7 & 92.0& 70.6 & 77.6& 60.7 & 67.9 & 83.3 & \textbf{92.0} & 80.4 & 94.3 \\
\bottomrule
\end{tabular}
}       
\setlength{\abovecaptionskip}{0.1cm}
\setlength{\belowcaptionskip}{0cm}
     \caption{Ablation studies. }
     \vspace{-12pt}
     \label{tab:ablation}
\end{table*}

Table~\ref{tab:main} shows the performance of the compared methods on GPT-3.5. 
We can observe the followings. 
1) $T^3$ outperforms all compared methods in AUROC and PRAUC on all datasets except HANS and PIQA, and in ECE on all datasets except SNLI, demonstrating its effectiveness. 
2) After combining $T^3$ with other methods \ie Top-$K$ Verb and PE, our method surpasses all compared methods on all datasets in the three evaluation metrics, showing the potential and flexibility of $T^3$ in combining with others to further improve self-detection. 
3) Hybrid with Top-$K$ Verb usually improves $T^3$'s performance in AUROC and PRAUC, which is in line with the performance improvement from Self-cons/CoT-cons to Hybrid. 
4) CAPE is very competitive in AUROC and PRAUC, showing that the self-detection is largely influenced by the prompt. Combining $T^3$ with PE usually improves $T^3$ in AUROC and PRAUC except for SNLI and Flipkart, which is in line with the performance decrease from Top-$K$ Verb to CAPE. This is potentially related to the prompt sensitivity of these methods and the specific prompts adopted. 
5) For other methods, CoT-cons outperforms Self-cons in AUROC and PRAUC in 5 out of 6 datasets, as many tasks performs better with CoT reasoning. P(True) has ambivalent results which limits its applicability.

\subsection{In-depth Analysis}

\paragraph{Ablation Studies.}\label{sec:ablation_studies}
We conduct the following ablation studies to further validate the rationality of our framework design.
1) \textbf{\emph{w/ CoT expl}}: substituting $e_1, ..., e^N$ in $p^v$ with $N$ different CoT reasoning generated from $p(q)$ to reveal the rationality of reflection on various answers. 
2) \textbf{\emph{sep expl}}: placing a single $e_i$ in $p^v$ each time and calculating the averaged detection score to reveal the effectiveness of joint considering all $e_i$ in one $p^v$. 
3) \textbf{\emph{w/o shuffle}}: ablating the order shuffling of $e_i$ in $p^v$. 

From Table~\ref{tab:ablation}, we can observe that: 
1) \emph{w/ CoT expl} largely underperforms $T^3$ on all three tasks, demonstrating the rationality of pushing LLM to reflect and justify from each answer's perspective. 
2) \emph{sep expl} underperforms $T^3$ on both SA and NLI tasks, showing that jointly considering multiple justifications in one prompt is often more beneficial, and thus we choose this setting. It slightly outperforms $T^3$ on the CQA task, potentially due to the higher independency and objectivity of the answer choices.
3) \emph{w/o shuffle} also underperforms $T^3$, indicating that there exists order sensitivity for $e_i$. Order shuffling and score average improve self-detection by mitigating their position bias.

\paragraph{Effect on Bias Mitigation. }

\begin{figure}[t!]
    \centering
    \setlength{\abovecaptionskip}{0.1cm}
\setlength{\belowcaptionskip}{0cm}
    \includegraphics[width = 0.49\textwidth]{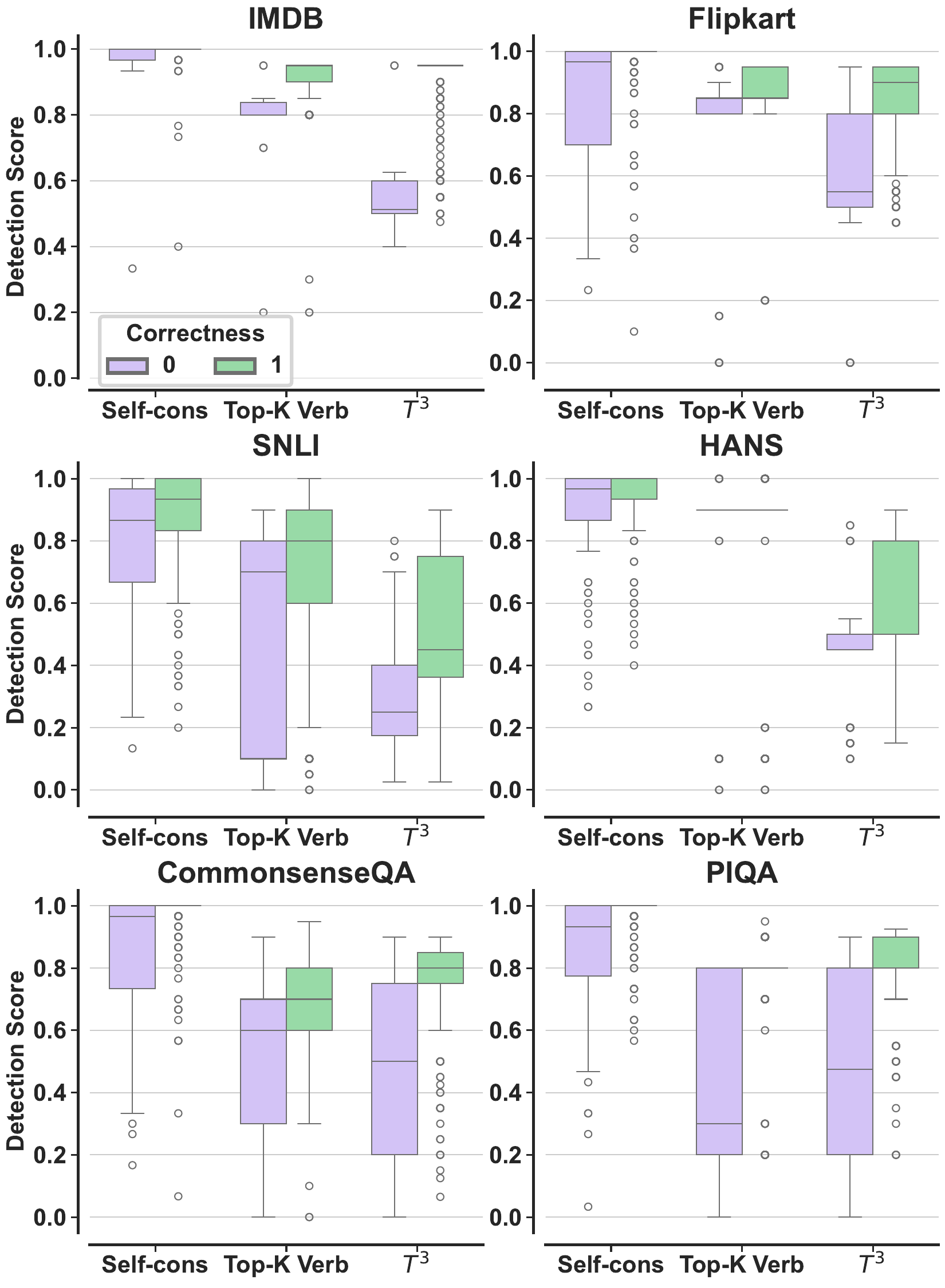}
    \caption{Visualization of bias mitigation effect of $T^3$ which largely reduces the detection score overlaps between correct (right) and incorrect (left) instances. }
    \label{fig:bias_mitigation}
    % \vspace{-20pt}
\end{figure}

Since our goal of improving self-detection is to reduce the over-trust on incorrect answers, we show the statistics of the detection scores for each dataset regarding the answer correctness in Figure~\ref{fig:bias_mitigation} to reveal the mechanism of $T^3$. 
We compare $T^3$ with Self-cons and Top-$K$ Verb which are witnessed with over-trust. 
We can observe that $T^3$ clearly reduces the detection score overlaps between correct and incorrect $q, a$ instances on all datasets, and significantly decreases the detection scores on incorrect instances in IMDB, Flipkart, SNLI and HANS. 
Thus, the answer accuracy is more separable by the detection score, achieving better self-detection. 

\begin{figure}[t!]
    \centering
    \setlength{\abovecaptionskip}{0.1cm}
\setlength{\belowcaptionskip}{0cm}
    \includegraphics[width = 0.33\textwidth]{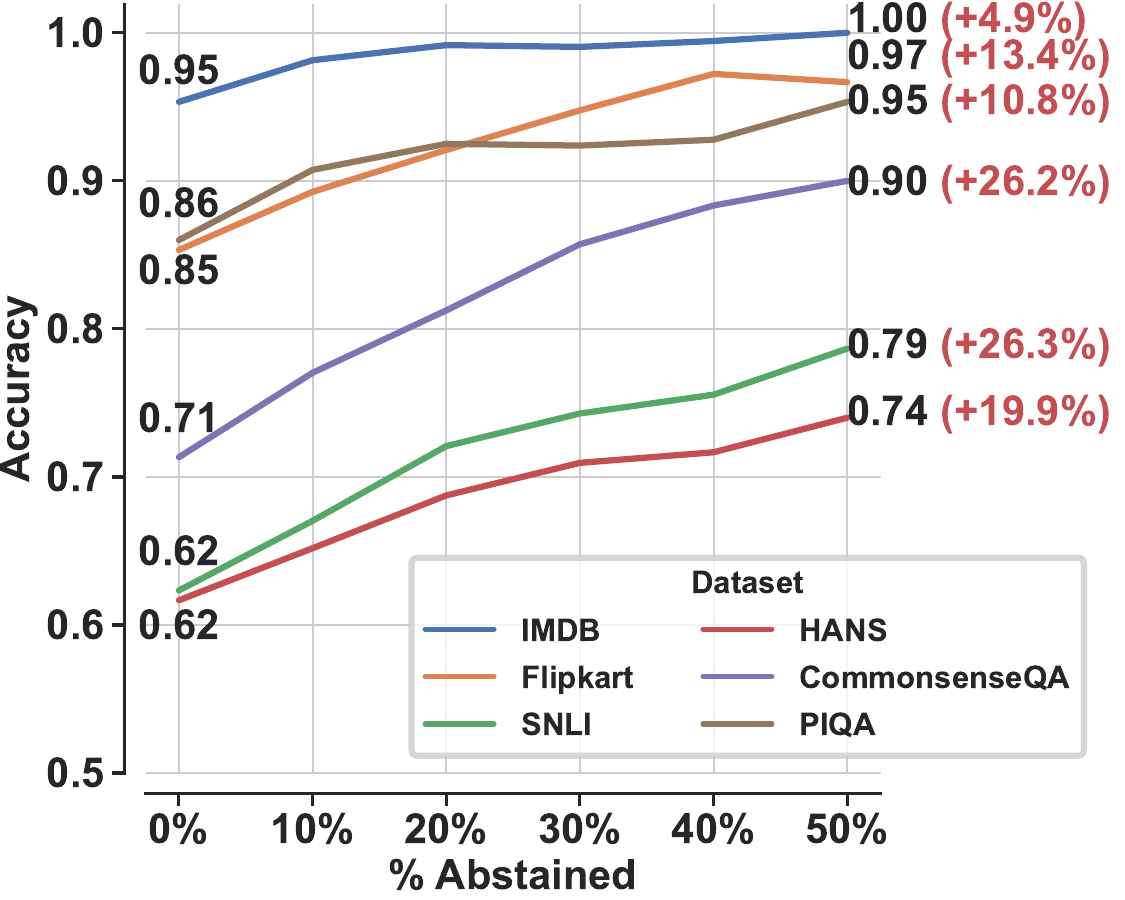}
    \caption{Accuracy improvement of selective prediction on $T^3$ detection scores. }
    \label{fig:selective_prediction}
    \vspace{-8pt}
\end{figure}

\paragraph{Effect on Selective Prediction via Detection Score.}
To show the utility of the detection score, we conduct experiments in selective prediction. 
The idea of selective prediction is to abstain the LLM-generated answers with low detection score to maintain better accuracy of the remaining instances. 
In Figure~\ref{fig:selective_prediction}, we show the accuracy of the remaining instances by abstaining 0\% - 50\% of instances with the lowest detection scores from $T^3$.
We can observe that by increasing the percentage of abstained instances, the accuracy for these datasets gradually improves around 10\% - 30\%, and IMDB even achieves 100\% accuracy. Naturally, the increase for datasets with lower accuracy is generally easier than datasets with higher accuracy. The result shows that $T^3$ possess strong potential to be applied in selective prediction scenarios.

\paragraph{Analysis on the Robustness of $T^3$.}
We evaluate the robustness of $T^3$ from three aspects: different target answers, different LLMs, and parameter sensitivity. In addition, we examine prompt sensitivity of $p^e$ and $p^v$ in Appendix~\ref{sec:appe_prompt_sensitivity}. 

Firstly, the generation of target answer $a$ may vary under LLM randomness, \eg setting the temperature greater than 0. 
We verify the robustness of $T^3$ by utilizing \textbf{different target answers}, \ie the majority answer of Self-cons ($a^{sc}$) and CoT-cons ($a^{cc}$), respectively, as shown in Table~\ref{tab:change_base_answer}.
We can observe the following. 
1) For both sets of target answers, $T^3$ largely outperforms baselines, showing its effectiveness.
2) Different target answers may have very different self-detection performance. 
Specifically, $a^{cc}$ on CommonsenseQA has a sharp decrease in AUROC of $T^3$ and CoT-cons compared with the other target answers, which is potentially due to the majority voting with CoT explanation diminished the the effect of the explanations in $T^3$.

\begin{table}[t!]
    \centering
   \resizebox{0.40\textwidth}{!}{
\begin{tabular}{lllll}
\toprule
                                           &            & Flipkart & HANS & CommonsenseQA \\
                                           \midrule 
\multirow{4}{*}{$a^{sc}$} & Self-cons  & 72.7     & 52.7 & 68.2 \\
                                           & CoT-cons   & 74.4     & 57.5 & 80.4 \\
                                           & Top-$K$ Verb & 80.4     & 51.8 & 69.2 \\
                                           & $T^3$       & \textbf{82.2}     & \textbf{69.5} & \textbf{82.7} \\
%\cmidrule{2-5}
% & Accuracy & 84.7 & 63.3 & 72.7\\
\midrule
\multirow{4}{*}{$a^{cc}$} & Self-cons  & 78.3     & 57.0 & 68.1 \\
                                           & CoT-cons   & 79.2     & 57.8 & 74.3 \\
                                           & Top-$K$ Verb & 83.9     & 53.3 & 67.5 \\
                                           & $T^3$       & \textbf{84.3}     & \textbf{69.2} & \textbf{75.0} \\
%\cmidrule{2-5}
% & Accuracy & 82.0 & 61.3 & 76.0\\
\bottomrule
\end{tabular} 
}
\setlength{\abovecaptionskip}{0.1cm}
\setlength{\belowcaptionskip}{0cm}
    \caption{AUROC on two different target answers. }
    \label{tab:change_base_answer}
    \vspace{-12pt}
\end{table}

Secondly, we evaluate $T^3$ on \textbf{different LLMs}. 
Table~\ref{tab:change_base_llm} shows the performance comparison of Flipkart, HANS and CommonsenseQA on GLM-4. 
We can observe that across different LLMs, combining $T^3$ with PE or Top-$K$ Verb
outperforms compared methods, validating its effectiveness. 
Besides, the self-detection ability may vary greatly across LLMs, \eg $T^3$'s AUROC of HANS on GLM-4 largely outperforms that on GPT-3.5. More results on Gemini can be found in Appendix~\ref{appe:different_LLMs}. 
% Moreover, Hybrid does not stably outperform single method. 

% \begin{table}[t!]
%     \centering
%     \resizebox{0.5\textwidth}{!}{
% \begin{tabular}{lllll}
% \toprule
%                  & GPT-4 & & GLM-4 &  \\
%                  % & ACC = 85.3 & ACC = 80.7 \\
%                 \cmidrule{2-3}\cmidrule{4-5} & AUROC & PRAUC & AUROC &PRAUC \\
% \midrule
% % CoT-cons         & 70.5  & 90.9& 73.4 & 88.8\\
% Top-$K$ Verb       & 80.8  & 94.3& 81.1 & 92.1 \\
% Hybrid           & 81.7  & 94.7& 80.4 & 92.0 \\
% \midrule
% TTA              & 81.0  & 94.5& \textbf{83.3} & \textbf{93.4}\\
% TTA + Top-$K$ Verb & \textbf{82.2} & \textbf{94.9}& 82.7 & 93.2\\
% \bottomrule
% \end{tabular}  
%     }
%     \setlength{\abovecaptionskip}{0.1cm}
% \setlength{\belowcaptionskip}{0cm}
%     \caption{Performance comparison of Flipkart on different LLMs. %The accuracy of the target answers is shown in the row header. 
%     }
%     \label{tab:change_base_llm}
%     \vspace{-5pt}
% \end{table}

\begin{table}[t!]
    \centering
    \resizebox{0.49\textwidth}{!}{
\begin{tabular}{lllllll}
\toprule
                 & \multicolumn{2}{l}{Flipkart}  & \multicolumn{2}{l}{HANS}      & \multicolumn{2}{l}{CommonsenseQA} \\
\cmidrule{2-3}\cmidrule{4-5}\cmidrule{6-7} & AUROC$\uparrow$ & PRAUC$\uparrow$ & AUROC$\uparrow$ & PRAUC$\uparrow$  & AUROC $\uparrow$ & PRAUC $\uparrow$ \\
\midrule
CoT Cons         & 73.4          & 88.8          & 66.4          & 87.5          & 83.1            & 97.0            \\
Top-$K$ Verb       & 81.1          & 92.1          & 65.4          & 88.0          & 72.3            & 95.3            \\
Hybrid           & 80.4          & 92.0          & 69.9          & 89.4          & 79.4            & \underline{97.2}      \\
CAPE             & 82.3          & 92.7          & \underline{82.4}    & \underline{94.0}    & \underline{80.0}      & 96.8            \\
\midrule
$T^3$             & \underline{83.3}    & \textbf{93.4} & 82.0          & 93.9          & 72.5            & 96.0            \\
\ \small{+ Top-$K$ Verb} & 82.7          & 93.2          & 80.9          & 93.9          & \textbf{81.0}   & \textbf{97.6}   \\
\ \small{+ PE}         & \textbf{83.8} & \textbf{93.4} & \textbf{84.9} & \textbf{95.7} & 76.9            & 96.6           \\
\bottomrule
\end{tabular}
    }
    \setlength{\abovecaptionskip}{0.1cm}
\setlength{\belowcaptionskip}{0cm}
    \caption{Performance comparison of Flipkart, HANS and CommonsenseQA on GLM-4. 
    }
    \label{tab:change_base_llm}
    \vspace{-5pt}
\end{table}

\begin{figure}[t!]
    \centering
    \setlength{\abovecaptionskip}{0.1cm}
\setlength{\belowcaptionskip}{0cm}
    \includegraphics[width = 0.48\textwidth]{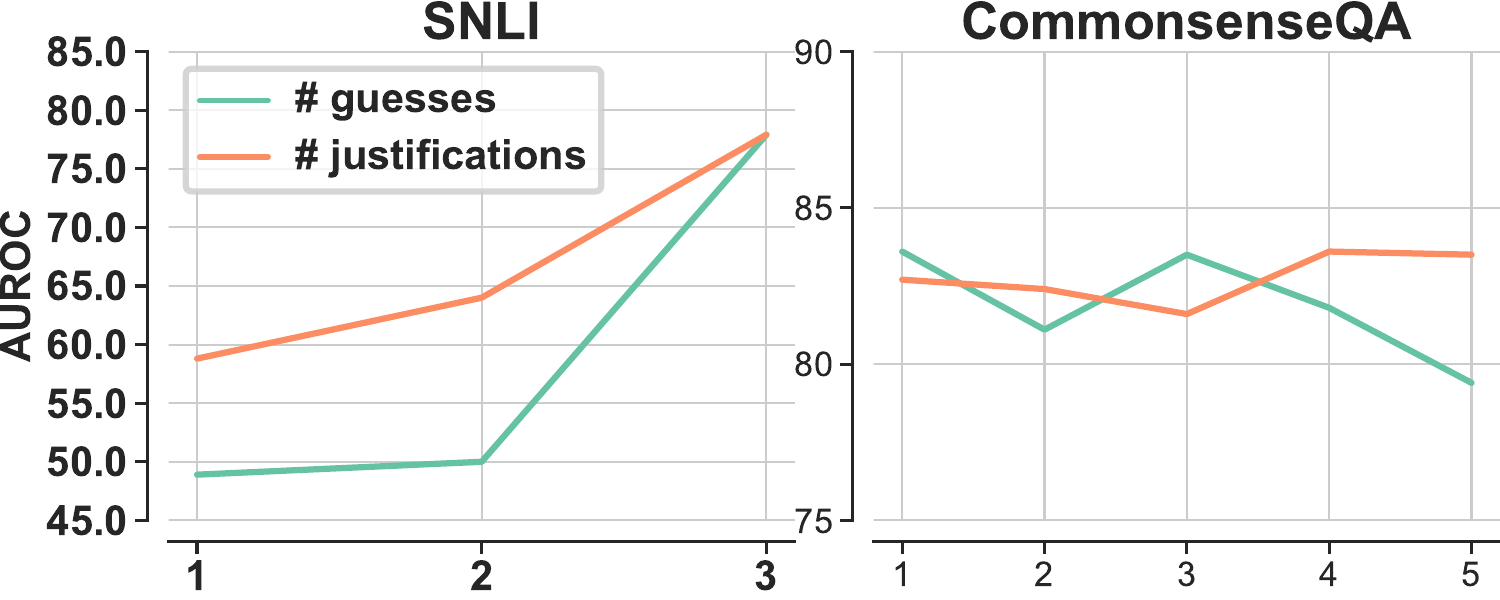}
    \caption{Parameter sensitivity, \ie changing the number of justifications and number of guesses in $p^v$. }
    \label{fig:param_sensitivity}
    \vspace{-8pt}
\end{figure}

Thirdly, we evaluate the \textbf{parameter sensitivity} of $T^3$ by changing the number of justifications and number of guesses in $p^v$. We conduct experiments on CommonsenseQA with five answer choices, and SNLI with three answer choices. 
From Figure~\ref{fig:param_sensitivity}, we can observe the followings.  
1) A larger number of justifications increases the performance on both datasets, indicating a sufficient number of justifications is vital for better self-detection.
2) Increasing the number of guesses results in a significant performance improvement on the SNLI dataset, revealing that enough number of guesses is demanded for the NLI task.
3) Comparably, the change in the number of guesses has a slight effect on the performance of the CommonsenseQA dataset, which is potentially because the CQA task is more objective than NLI.

\section{Conclusion}

In this paper, we tackled the over-trust issue of self-detection on black-box API LLMs. 
We categorized existing methods into two paradigms and pointed out their limitation of merely evaluating on LLM-generated answer with potential LLM over-trust. We proposed a novel paradigm to address this limitation by comprehensively evaluating the trustworthiness of multiple candidate answers in the answer space. 
Following our paradigm, we presented a two-step framework $T^3$ by asking LLM to reflect and justify the validity of each answer for joint confidence calibration. Our framework achieved improved self-detection performance over compared methods and was combined with existing methods for further improvement. % 需要说出具体是什么指标吗？
In future work, we will explore the combination of $T^3$ with more methods, and its utility in white-box LLMs. 
\clearpage
\newpage
\section*{Limitations}
Our work has several limitations. 
Firstly, our research scope is limited to the self-detection for black-box API LLM. 
While our framework is suitable for many state-of-the-art LLMs in this form, it might not be optimal for white-box LLMs, which offer access to more model information, thus limiting its broader applicability.
Secondly, the utility of self-detection is not primarily studies in this work. 
Although we demonstrate the utility of detection scores in selective prediction scenarios,
the challenge still lies in leveraging them to enhance task accuracy or enable LLM self-correction, calling for further exploration. 
Lastly, our framework lacks consideration in prompt optimization for self-detection, an area where future self-detection methods are expected to consider.
% , since various LLMs have diverse prompt preferences. 

\section*{Ethics Statement}
Our ethical concerns involve the following. 
First, our experimental results are mainly obtained in English datasets, where the applicability on other languages are not comprehensively evaluated. 
Secondly, our research scope is black-box API LLMs, where open-sourced LLMs are more advocated for its reproducibility. 
Finally, the self-detection of LLM may mislead people to blindly trust LLM and easily accept untrustable answers, causing potential harms. 
% Scientific work published at ACL 2023 must comply with the ACL Ethics Policy.\footnote{\url{https://www.aclweb.org/portal/content/acl-code-ethics}} We encourage all authors to include an explicit ethics statement on the broader impact of the work, or other ethical considerations after the conclusion but before the references. The ethics statement will not count toward the page limit (8 pages for long, 4 pages for short papers).

\section*{Acknowledgements}

This work is supported by the NExT Research Center. We thank the reviewers for their constructive feedback.

\bibliography{anthology,custom}
\bibliographystyle{acl_natbib}

\appendix \label{sec:appendix}

\section{Details for compared methods.} \label{sec:appe_detail}

\subsection{LLM Hyperparameters.} \label{sec:appe_llm_param}

For all LLMs, we set the maximum token as 200. 
For GPT-3.5 and Gemini, if sampling a single response ($N=1$), we set the temperature as 0, and other hyperparameters as default. If sampling multiple responses, we sample $N=30$ ($N=5$ for Gemini due to API call limitation) responses with temperature as 1, which is only for Self-cons, CoT-cons, and P(True). Specially, for Self-detect we sample 15 rephrasing for each question with temperature as 1, and one answer for each rephrased question with temperature as 0, following the original paper. 
For GLM-4, if sampling a single response, we set the do\_sample as False. If sampling a variety of responses, we set temperature as 0.9, top p as 0.9, and $N=5$. 
Note that these LLM hyperparameters are not carefully tuned. 

\subsection{Dataset Detail.} \label{sec:appe_dataset}
Due to the cost limitation, we randomly sample 300 training data for each dataset in our experiments. 
For IMDB and SNLI datasets, we use the same randomly sampled 300 data sets as the CAD SA and NLI in the preliminary experiments. 
We will release the dataset splits. 
Table~\ref{tab:answer_choices} shows the number and examples of candidate answers for each dataset. 

\begin{table}[]
    \centering
    \resizebox{0.5\textwidth}{!}{
\begin{tabular}{lll}
\toprule
              & $N$ & examples                           \\
\midrule
IMDB          & 2 & positive negative                  \\
Flipkart      & 2 & positive negative                  \\
SNLI          & 3 & entailment, neutral, contradiction \\
HANS          & 2 & entailment, non entailment         \\
CommonsenseQA & 5 & (a) yard, (b) basement, \\
&&(c) kitchen, (d) living room, (e) garden          \\
PIQA          & 2 & (a) pour it onto a plate, (b) pour it into a jar                          \\
\bottomrule
\end{tabular}
}
    \caption{The number ($N$) and examples of candidate answers for each dataset.}
    \label{tab:answer_choices}
\end{table}

\subsection{Prompts} \label{sec:appe_prompts}
The basic instructions for different datasets are shown as below, where [] refers to specific task inputs.
\begin{itemize}
    \item \textbf{IMDB}: \\
          \textit{Given a piece of movie review, classify the attitude to the movie as Positive or Negative. [text]}
    \item \textbf{Flipkart}: \\
          \textit{Given a piece of text, classify the sentiment as Positive or Negative. [text] }
    \item \textbf{SNLI}: \\
          \textit{Determine whether the hypothesis is an entailment (can be logically inferred from the premise), a contradiction (cannot be true given the premise), or neutral (does not have enough information to determine its truth value). Premise: [premise] Hypothesis: [hypopthesis]. }
    \item \textbf{HANS}: \\
          \textit{Determine whether the second sentence in each pair logically follows from the first sentence. The output is either "entailment" if the second sentence logically follows from the first, or "not entailment" if it does not. Sentence 1: [sentence1]. Sentence 2: [sentence2].}
    \item \textbf{CommonsenseQA}: \\
          \textit{Read the given question and select the most appropriate answer by indicating the associated letter. Question: [question]. Answer choices: (a) $a^q_1$ (b) $a^q_2$ (c) $a^q_3$ (d) $a^q_4$ (e) $a^q_5$. }
    \item \textbf{PIQA}: \\
          \textit{Read the given question and select the most appropriate answer by indicating the associated letter. Question: [question]. Answer choices: (a) $a^q_1$ (b) $a^q_2$.}
\end{itemize}

The prompts for compared methods are shown below, where \emph{[instruction]} denotes the task instruction with the task input, and \emph{[instruction\_only]} denotes the instruction without task input.

\begin{itemize}
    \item \textbf{Self-cons}: \textit{[instruction].}
    
    \item \textbf{CoT-cons}: \\
    \textit{[instruction]. Please output strictly following this format: Explanation: [reasons for the sentiment label] Answer: [Positive or Negative]}
    
    \item \textbf{Top-$K$ Verb $p^b$}: \\
    \textit{The task is to [instruction\_only]. Provide your n best guesses and the probability that each is correct (0.0 to 1.0) for the following question. Give ONLY the guesses and probabilities, no other words or explanation. For example:\\
    G1: <first most likely guess, as short as possible; not a complete sentence, just the guess!>\\
    P1: <the probability between 0.0 and 1.0 that G1 is correct, without any extra commentary whatsoever; just the probability!> ... GN: <N-th most likely guess, as short as possible; not a complete sentence, just the guess!>\\
    PN: <the probability between 0.0 and 1.0 that GN is correct, without any extra commentary whatsoever; just the probability!> [question] [answer choices]. }
    
    \item \textbf{P(True) $p^t$}:\\
    \textit{The task is to [instruction]. Label: [label]. Is the label correct or incorrect?}
    
    \item \textbf{Self-detect}: \\
    For question rephrasing: \textit{Paraphrase the given sentence. Please make sure the paraphrased sentence has exactly the same meaning as the original sentence. [question] } \\
    For inference: \textit{[instruction].}
    
    \item \textbf{CAPE}: \\
    \textit{
    Provide your 2 best guesses and the probability that each is correct (0.0 to 1.0) for the following task. Give ONLY the guesses and probabilities, no other words or explanation. For example:\\
    G1: <first most likely guess, as short as possible; not a complete sentence, just the guess!>\\
    P1: <the probability between 0.0 and 1.0 that G1 is correct, without any extra commentary whatsoever; just the probability!> ... GN: <N-th most likely guess, as short as possible; not a complete sentence, just the guess!>\\
    PN: <the probability between 0.0 and 1.0 that GN is correct, without any extra commentary whatsoever; just the probability!> 
    Instruction: [instruction\_only] [question] \\
     - $a^q_1$ (or A. $a^q_1$)\\
     ... \\
     - $a^q_N$ (or N. $a^q_N$) \\
     Possible explanation 1: [$e^1$] \\
     ... \\
     Possible explanation N: [$e^N$] \\
     Correct Choice:}\\

    \item \textbf{$T^3$ $p^v$}: \\
    \textit{The task is to [instruction\_only]. Provide your n best guesses and the probability that each is correct (0.0 to 1.0) for the following question. Give ONLY the guesses and probabilities, no other words or explanation. For example:\\
    G1: <first most likely guess, as short as possible; not a complete sentence, just the guess!>\\
    P1: <the probability between 0.0 and 1.0 that G1 is correct, without any extra commentary whatsoever; just the probability!> ... GN: <N-th most likely guess, as short as possible; not a complete sentence, just the guess!>\\
    PN: <the probability between 0.0 and 1.0 that GN is correct, without any extra commentary whatsoever; just the probability!> \\
    \text{[question]} [answer choices]. \\
    Possible explanation 1: [explanation 1]. \\
    ...\\
    Possible explanation N: [explanation N].}
    
\end{itemize}

\subsection{Additional Implementation Detail. } \label{sec:appe_add_detail}

For $T^3$ and Top-$K$ Verb, the $N$ is set to the number of candidate answers for each dataset as in Table~\ref{tab:answer_choices}. 

For the shuffling of the justification order in $p^v$, we use one original and one reversed order for $T^3$ on all datasets. For datasets with more than two justifications (SNLI and CommonsenseQA), we set the original justification order for SNLI as ``entailment, neutral, contradiction" and follow the given answer choice order for CommonsenseQA in the dataset. 

CAPE is prompt ensemble for Top-$K$ Verb. We follow the original paper to adopt two multi-choice template with alphabetic or itemized labels in addition to the original Top-$K$ Verb prompt (See Section~\ref{sec:appe_prompts}). For each multi-choice template, we use the original and the reversed label orders. In total, the confidence score is an average of five prompts. 

For $T^3$ + PE, we put $T^3$ into the multi-choice template with alphabetic labels, and use two reversed label orders and 2 reversed justification orders, in total four prompts. 

The number of API calls for different methods are shown in Table~\ref{tab:api_call}. We can observe that compared with other methods $T^3$ does not incur large increase in number of calls. In our experiments, the maximum value of $N$ is 5. Considering its effectiveness, the cost of $T^3$ is reasonable.  

% number of API calls.
\begin{table}[hbt!]
    \centering
    \resizebox{0.5\textwidth}{!}{
\begin{tabular}{llllll}
\toprule
           & Self-cons   & CoT-cons & Top-$K$ Verb & P(True)          & Hybrid   \\
\midrule
\# call & 30          & 30       & 1          & 30               & 31       \\
\midrule
           & Self-detect & CAPE     & $T^3$        & $T^3$ + Top-$K$ Verb & $T^3$ + PE \\
\midrule
\# call & 30          & 5        & N+2        & N+3              & N+4     \\
\bottomrule
\end{tabular}
    }
    \caption{Comparison on the number of API calls of compared methods, where $N$ denotes the number of choices for different datasets. }
    \label{tab:api_call}
\end{table}

\section{Implementation Detail for Preliminary Experiments.} \label{sec:appe_prelim_exp}
For the preliminary experiments, we randomly sample 300 instances from the training set of CAD SA and NLI, respectively. 
For those original questions with more than one counterfactual questions, we randomly select one counterfactual question for experiment. 
The prompts can be viewed in Section~\ref{sec:appe_prompts}. 
CAD SA is annotated from IMDB, and CAD NLI is annotated from SNLI. 
The \emph{w/ cf} is based on Top-$K$ Verb, which is better calibrated than Self-cons. 
For \emph{w/ cf}, we obtain the Top-$K$ Verb outputs for counterfactual and original questions, respectively. 
We use the guess with the largest probability in the response as the answer to $\bar{q}$, and the probability as its confidence score. 
The LLM is GPT-3.5 (\emph{gpt-3.5-1106}). See Section~\ref{sec:appe_llm_param} for LLM hyperparameters. 

% Additional results for \emph{w/ cf} can be found in Table. 
% additional results for prelim exp. including auroc, acc?

\begin{table}[hbt!]
    \centering
    \resizebox{0.35\textwidth}{!}{
\begin{tabular}{llll}
\toprule
  & PIQA       & HANS       & Flipkart   \\
\midrule
$p^e$ & 84.2 $\pm$ 2.0 & 62.7 $\pm$ 4.3 & 78.0 $\pm$ 2.2 \\
$p^v$ & 83.0 $\pm$ 0.5 & 68.3 $\pm$ 1.7 & 81.2 $\pm$ 0.3 \\
\bottomrule
\end{tabular}
    }
    \caption{The average and standard deviation of AUROC for $T^3$ with different rephrasing of prompts on GPT-3.5. }
    \label{tab:prompt_sensitivity}
    \vspace{-15pt}
\end{table}

\section{Prompt Sensitivity} \label{sec:appe_prompt_sensitivity}

We examine the prompt sensitivity of $p^e$ and $p^v$ by rephrasing each of them three times with ChatGPT\footnote{\url{https://chat.openai.com/}.} and compute the average and standard deviation of AUROC, as shown in Table~\ref{tab:prompt_sensitivity}. We can observe the followings.
1) The variation of prompts has a mild effect on the performance  of $T^3$. Across the three datasets, HANS is the most sensitive to prompt rephrasing, potentially related to its lower AUROC performance. 
2) The change of $p^e$ has larger impact on the detection performance than $p^v$. This is probably because the justifications generated by $p^e$ have a larger space of variation than the outputs of $p^v$, \ie guesses and probabilities.

\begin{table}[ht!]
    \centering
    \resizebox{0.5\textwidth}{!}{
\begin{tabular}{lllllll}
\toprule
             & \multicolumn{2}{l}{Flipkart}  & \multicolumn{2}{l}{PIQA} & \multicolumn{2}{l}{CommonsenseQA} \\
\cmidrule{2-3} \cmidrule{4-5} \cmidrule{6-7} & AUROC         & PRAUC         & AUROC       & PRAUC      & AUROC           & PRAUC           \\
\midrule
CoT-cons     & 67.6          & 89.1          & 85.8          & 96.4          & 72.2            & 86.0            \\
Top-K Verb   & \underline{84.0}    & \underline{94.9}    & 64.7          & 91.8          & 67.0            & 83.9            \\
Hybrid       & \textbf{85.0} & \textbf{95.5} & 77.0          & 96.1          & 74.5            & 87.9            \\
CAPE         & 81.5          & 94.5          & \underline{86.5}    & \underline{97.8}    & \underline{77.3}      & \underline{89.5}      \\
\midrule
$T^3$           & 82.0          & 93.9          & 82.6          & 96.5          & 73.4            & 89.3            \\
+ Top-K Verb & 83.1          & 94.6          & 77.2          & 96.1          & 76.3            & \textbf{91.2}   \\
+ CAPE       & 74.8          & 92.1          & \textbf{90.5} & \textbf{98.3} & \textbf{77.5}   & 91.1          \\ 
\bottomrule
\end{tabular}
    }
    \setlength{\abovecaptionskip}{0.1cm}
\setlength{\belowcaptionskip}{0cm}
    \caption{Performance comparison of Gemini on Flipkart, PIQA and CommonsenseQA. 
    }
    \label{tab:gemini}
    \vspace{-5pt}
\end{table}

% \begin{table}[ht!]
%     \centering
%     \resizebox{0.35\textwidth}{!}{
% \begin{tabular}{lllll}
% \toprule
%                  % & GPT-4 & & GLM-4 &  \\
%                  % & ACC = 85.3 & ACC = 80.7 \\
%                 & AUROC & PRAUC \\
% \midrule
% % CoT-cons         & 70.5  & 90.9\\
% Top-$K$ Verb       & 80.8  & 94.3\\
% Hybrid           & 81.7  & 94.7 \\
% \midrule
% $T^3$              & 81.0  & 94.5\\
% + Top-$K$ Verb & \textbf{82.2} & \textbf{94.9}\\
% \bottomrule
% \end{tabular}  
%     }
%     \setlength{\abovecaptionskip}{0.1cm}
% \setlength{\belowcaptionskip}{0cm}
%     \caption{Performance of GPT-4 on Flipkart's top performing baselines. 
%     }
%     \label{tab:gpt-4}
%     \vspace{-5pt}
% \end{table}

\section{Additional Results on Different LLMs} \label{appe:different_LLMs}

In addition to GPT-3.5 and GLM-4, we show the results of Gemini on three datasets. From Table~\ref{tab:gemini}, we can observe that although $T^3$ outperforms all compared methods in PIQA and CommonsenseQA, it does not outperform all compared methods on Flipkart. 
By analyzing the outputs, we discover that Gemini cannot always follow the instruction to perform reflection and generated justification for the designated answer. Instead, it tends to perform answer prediction and followed by an explanation on its predicted answer. Without effective reflection and justification from different answers' perspectives, the effectiveness of $T^3$ is diminished. Therefore, the effectiveness of $T^3$ depends on the ability of the specific LLM in following the instructions in Table~\ref{tab:prompt_e_and_v}. 
% Additionally, the results on GPT-4 (\cf~\ref{tab:gpt-4}) also show that $T^3$ outperforms top-performing compared methods on Flipkart. Due to cost constraints, we do not conduct comprehensive experiments on GPT-4. 

\section{Additional Baselines. }
We present the results of more baselines on GPT-3.5, including the white-box average token probability \cite{manakul-etal-2023-selfcheckgpt}, and the black-box induced consistency and self-probing \cite{xiong2023can}. 
From Table~\ref{tab:white-box}, we can observe that the white-box method largely underperforms the black-box baselines across three tasks, indicating its limited capability on large LLMs with strong instruction-following abilities.

\begin{table}[ht!]
    \centering
    \resizebox{0.5\textwidth}{!}{
\begin{tabular}{lllllll}
\toprule
             & \multicolumn{2}{l}{Flipkart}  & \multicolumn{2}{l}{HANS}      & \multicolumn{2}{l}{CommonsenseQA} \\
             & AUROC         & PRAUC         & AUROC         & PRAUC         & AUROC           & PRAUC           \\
\midrule
Avg Tok Prob & 32.4          & 78.0          & 56.0          & 65.5          & 43.3            & 67.3            \\
Induced-cos   & 64.5          & 89.2       & 63.4          & 71.0         & 71.4           & 81.6            \\
Self-Probe   & 72.8          & 92.3         & 58.2         & 66.1          & 58.6            & 75.1           \\
$T^3$            & \textbf{81.3} & \textbf{94.5} & \textbf{69.9} & \textbf{77.5} & \textbf{83.5}   & \textbf{90.7}  \\
\bottomrule
\end{tabular}
    }
    \setlength{\abovecaptionskip}{0.1cm}
\setlength{\belowcaptionskip}{0cm}
    \caption{Performance comparison of additional white-box and black-box baselines.  
    }
    \label{tab:white-box}
    \vspace{-5pt}
\end{table}

\section{Accuracy Statistics}
We report the accuracy for each dataset in our experiments, and the precision, recall and F1 with $T^3$ confidence threshold of 0.5 (0.25 for SNLI), as shown in Table~\ref{tab:acc_stat}. 
It is clear that the F1 value positively correlates to the accuracy. And the recall is generally higher than precision, showing that LLM still tends to over-trust the answer.

\begin{table}[h]
\centering
\resizebox{0.5\textwidth}{!}{
\begin{tabular}{lllll}
\toprule
Dataset        & Accuracy & Precision & Recall & F1   \\
\midrule
IMDB          & 0.95     & 0.96      & 0.99   & 0.98 \\
Flipkart      & 0.85     & 0.88      & 0.98   & 0.93 \\
SNLI          & 0.62     & 0.74      & 0.91   & 0.82 \\
HANS          & 0.62     & 0.69      & 0.89   & 0.78 \\
CommonsenseQA & 0.71     & 0.81      & 0.92   & 0.86 \\
PIQA          & 0.86     & 0.94      & 0.97   & 0.92 \\
\bottomrule
\end{tabular}
}
    \caption{The accuracy for each dataset, and the precision, recall, F1 of $T^3$. }
    \label{tab:acc_stat}
\end{table}

\section{Results on Open-Ended QA} 
In addition to the multi-choice QA tasks where candidate answers are given, we also conduct experiments on the open-ended QA task of TruthfulQA \cite{lin2022truthfulqa}. 
Since the open-ended QA has no candidate answer, we employ model ensemble to generate candidate answers and form the answer space. 
Specifically, for each question, we start by generating one answer from GPT-3.5, and then use an outstanding smaller LM GLM-4, to sample one additional answer, forming the answer space of two answers. We believe the answer space contains more plausible answers than those produced by GPT-3.5 alone, thus allowing GPT-3.5 to reflect on its generated answers and assign more calibrated confidence to them.
We utilize GPT-4 (\emph{gpt-4-0613}) to annotate the factuality of the generated answers. 
Performance of several strong baselines are shown in Table~\ref{tab:open-ended}, demonstrating the outstanding self-detection ability of $T^3$ in open-ended QA.  
We can also observe that Hybrid with other method may be an effective approach in improving the self-detection performance, and the prompt ensemble of CAPE does not seem very effective in this specific scenario, requiring further prompt optimization. 

\begin{table}[h]
    \centering
\resizebox{0.45\textwidth}{!}{
\begin{tabular}{llll}
\toprule
TruthfulQA      & AUROC    & PRUAC    & ECE      \\
\midrule
Self-cons       & 61.1     & 73.8     & 0.096    \\
CoT-cons        & 62.9     & \textbf{76.1}     & 0.133    \\
Top-$K$ Verb    & 61.8     & 74.9     & \underline{0.056}    \\
Hybrid          & \underline{63.0}    & 75.0     & \textbf{0.051}    \\
CAPE            & 54.7     & 72.6     & 0.156    \\
$T^3$           & 62.3     & 75.2     & 0.060    \\
$T^3$ + Top-$K$ Verb & \textbf{64.4}     & \underline{76.0}     & 0.078    \\
\bottomrule
\end{tabular}
}
    \caption{Performance comparison on open-ended QA task of TruthfulQA.}
    \label{tab:open-ended}
\end{table}

\section{Case study} \label{appe:case_study}
We present two case study of PIQA. 
From Table~\ref{tab:case_study}, we can observe that the detection score of the incorrect answer (a) is lowered by $T^3$ (0.7 $\to$ 0.45). The justification (b) points out the reason why (b) is the preferred answer, which increased the confidence in (b) and in turn decreases (a)'s detection score. 
From Table~\ref{tab:case_study_2}, we can observe that the LLM is not sure whether the sponge should be dampened, thus having an ambivalent confidence score of 0.5 is better than being 0.7 confident on (b). 

\begin{table}[hbt!]
    \centering
    \resizebox{0.5\textwidth}{!}{
\begin{tabular}{p{8cm}}
\toprule
\textbf{Input}: The task is to read the given question and select the most appropriate answer by indicating the associated letter. Provide your 2 best guesses and the probability that each is correct (0.0 to 1.0) for the following question. Give ONLY the guesses and probabilities, no other words or explanation. For example:

G1: <first most likely guess, as short as possible; not a complete sentence, just the guess!>

P1: <the probability between 0.0 and 1.0 that G1 is correct, without any extra commentary whatsoever; just the probability!> ... G2: <2-th most likely guess, as short as possible; not a complete sentence, just the guess!>

P2: <the probability between 0.0 and 1.0 that G2 is correct, without any extra commentary whatsoever; just the probability!> \\ 

Question: How do you repair a torn shirt?\\
Answer Choices: \\
(a) To repair a torn shirt, prepare the needle with a piece of thread tied to it. Pull together the fabric where the shirt is torn and sew together. \\
(b) Prepare the needle with a piece of thread tied to it. \textcolor{blue}{Flip the shirt inside-out}, pull together the fabric where the shirt is torn and sew together. \\
\textbf{Prediction: (a). Ground truth: (b).} \\
\midrule
\textbf{Top-$K$ Verb Output}: G1: A  P1: 0.7  G2: B  P2: 0.3. \textbf{Confidence}: \textcolor{blue}{0.7}  \\
\midrule
\textbf{Justification for (a)}:  The most appropriate answer is (a) because it provides a clear and direct explanation of how to repair a torn shirt. It instructs to prepare the needle with a piece of thread tied to it and then pull together the fabric where the shirt is torn and sew it together. This answer is straightforward and does not include unnecessary steps, making it the most appropriate choice. \\
\textbf{$p^v$ output 1}: G1: A  P1: 0.6  G2: B  P2: 0.4   \\
\textbf{Justification for (b)}: Option (b) is the most appropriate answer because \textcolor{blue}{it provides a more detailed and accurate description} of the process of repairing a torn shirt. \textcolor{blue}{It includes the step of flipping the shirt inside-out}, which is an important step in order to hide the stitches and make the repair less visible. This additional detail makes option (b) the better choice for repairing a torn shirt. \\
\textbf{$p^v$ output 2}: G1: B  P1: 0.7 G2: A  P2: 0.3 \\
\textbf{$T^3$ Detection Score}: (0.6 + 0.3)/2 = \textcolor{blue}{0.45} \\
\bottomrule
\end{tabular}
}
    \caption{Case study for PIQA. $p^v$ output 1 refers to $p^v$ with explanation (a) before explanation (b), and $p^v$ output 2 refers to the reversed order. }
    \label{tab:case_study}
\end{table}

\begin{table}[hbt!]
    \centering
    \resizebox{0.5\textwidth}{!}{
\begin{tabular}{p{8cm}}
\toprule
\textbf{Input}: The task is to read the given question and select the most appropriate answer by indicating the associated letter. Provide your 2 best guesses and the probability that each is correct (0.0 to 1.0) for the following question. Give ONLY the guesses and probabilities, no other words or explanation. For example:

G1: <first most likely guess, as short as possible; not a complete sentence, just the guess!>

P1: <the probability between 0.0 and 1.0 that G1 is correct, without any extra commentary whatsoever; just the probability!> ... G2: <2-th most likely guess, as short as possible; not a complete sentence, just the guess!>

P2: <the probability between 0.0 and 1.0 that G2 is correct, without any extra commentary whatsoever; just the probability!> \\ 

Question: How do I keep my favorite couch fur free?\\
Answer Choices: \\
(a) Purchase a dry sponge at a pet supply store and use it to lightly wipe dog hair in a sweeping motion from upholstery. \\
(b) Purchase a dry sponge at a pet supply store and \textcolor{blue}{dampen it} and use it to lightly wipe dog hair in a sweeping motion from upholstery. \\
\textbf{Prediction: (b). Ground truth: (a).} \\
\midrule
\textbf{Top-$K$ Verb Output}: G1: B  P1: 0.7  G2: A  P2: 0.3. \textbf{Confidence}: \textcolor{blue}{0.7}  \\
\midrule
\textbf{Justification for (a)}:  The most appropriate answer is (a) because using a dry sponge to wipe dog hair in a sweeping motion from upholstery will help to remove the fur without getting the couch wet. \textcolor{blue}{Dampening the sponge, as suggested in option (b), may lead to the couch getting wet and potentially causing damage to the upholstery.} Therefore, option (a) is the best choice for keeping the favorite couch fur free. \\
\textbf{$p^v$ output 1}: G1: A  P1: 0.7  G2: B  P2: 0.3   \\
\textbf{Justification for (b)}: The most appropriate answer is (b) because \textcolor{blue}{dampening the dry sponge will help to better pick up and remove the dog hair from the upholstery.} Using a dry sponge alone may not be as effective in removing the fur. Therefore, dampening the sponge will provide better results in keeping the favorite couch fur free. \\
\textbf{$p^v$ output 2}: G1: B  P1: 0.7 G2: A  P2: 0.3 \\
\textbf{$T^3$ Detection Score}: (0.3 + 0.7)/2 = \textcolor{blue}{0.5} \\
\bottomrule
\end{tabular}
}
    \caption{Case study for PIQA. $p^v$ output 1 refers to $p^v$ with justification (a) before justification (b), and $p^v$ output 2 refers to the reversed order. }
    \label{tab:case_study_2}
\end{table}

\end{document}